\documentclass[conference]{IEEEtran}
\IEEEoverridecommandlockouts
\usepackage{cite}
\usepackage{amsmath,amssymb,amsfonts}
\usepackage{algorithmic}
\usepackage{graphicx}
\usepackage{textcomp}
\usepackage{xcolor}
\usepackage{cite}
\def\BibTeX{{\rm B\kern-.05em{\sc i\kern-.025em b}\kern-.08em
    T\kern-.1667em\lower.7ex\hbox{E}\kern-.125emX}}
\begin{document}

\title{AMaizeD: An End to End Pipeline for Automatic Maize Disease Detection
\thanks{IITI DRISHTI CPS Foundation.}
}

\author{\IEEEauthorblockN{Pawan K. Ajmera}
\IEEEauthorblockA{\textit{EEE Department} \\
\textit{BITS Pilani}\\
Pilani, India \\
pawan.ajmera@pilani.bits-pilani.ac.in}
\and
\IEEEauthorblockN{Sanchit M. Kabra}
\IEEEauthorblockA{\textit{CSIS Department} \\
\textit{BITS Pilani}\\
Pilani, India \\
kabrasanchit@gmail.com}

\and
\IEEEauthorblockN{Anish Mall}
\IEEEauthorblockA{\textit{EEE Department} \\
\textit{BITS Pilani}\\
Pilani, India \\
21.anishmall@gmail.com}
\and
\IEEEauthorblockN{Ankur Lhila}
\IEEEauthorblockA{\textit{EEE Department} \\
\textit{BITS Pilani}\\
Pilani, India \\
ankur.lhila@gmail.com}
}

\maketitle

\begin{abstract}
This research paper presents AMaizeD: An End to End Pipeline for Automatic Maize Disease Detection, an automated framework for early detection of diseases in maize crops using multispectral imagery obtained from drones. We also develop a custom hand-collected dataset focusing specifically on maize crops was meticulously gathered by expert researchers and agronomists. The dataset encompasses a diverse range of maize varieties, cultivation practices, and environmental conditions, capturing various stages of maize growth and disease progression. By leveraging multispectral imagery, the framework benefits from improved spectral resolution and increased sensitivity to subtle changes in plant health. The proposed framework employs a combination of convolutional neural networks (CNNs) as feature extractors and segmentation techniques to identify both the maize plants and their associated diseases. Experimental results demonstrate the effectiveness of the framework in detecting a range of maize diseases, including common rust, grey leaf spot and leaf blight. The framework achieves state-of-the-art performance on the custom hand-collected dataset and contributes to the field of automated disease detection in agriculture, offering a practical solution for early identification of diseases in maize crops using advanced machine learning techniques and deep learning architectures.
\end{abstract}

\begin{IEEEkeywords}
Machine Learning, Precision Agriculture
\end{IEEEkeywords}

\section{Introduction}
Maize holds significant agricultural value in India. With a production of approximately 28 million metric tons in the 2020-2021 period, this crop is among the most extensively cultivated in India \cite{maize_production}. Maize holds significant importance in the agricultural landscape of India, serving as a source of sustenance, animal fodder, and revenue for cultivators, thereby bolstering the nation's economy. The production quantities of these maize crops are affected by the presence of pests and diseases. This paper uses a custom maize dataset, and only the diseases included in this dataset are used. These include Blight, Common Rust and Gray Leaf Spot. Healthy images included in the dataset have also been used in this work.

In order to mitigate the impact of diseases and pests on crops, the agricultural industry has to employ expensive techniques and a range of pesticides \cite{pesticides_impact}. The unsolicited utilization of chemical methods on a large scale has detrimental effects on both plant and human health, in addition to causing adverse impacts on the environment. Additionally, these methods result in an escalation of production expenses. Image processing, however, is a method that can be employed in precision agriculture to identify the areas of infestation and minimize the use of pesticides \cite{image_processing}.Deep learning, a subfield of machine learning, has emerged as a powerful technique for automated image analysis and pattern recognition tasks. Convolutional Neural Networks (CNNs) have shown remarkable success in various computer vision applications, including object detection and classification. In the context of maize disease detection, CNNs have the potential to learn discriminative features directly from raw images, enabling accurate and automated disease identification. By leveraging the hierarchical structure of CNNs\cite{cnn}, lower-level features such as edges and textures can be learned in the initial layers, while higher-level features related to specific disease symptoms can be learned in the deeper layers. This hierarchical feature extraction allows the network to capture both local and global patterns, facilitating robust disease detection even in the presence of variations in illumination, scale, and orientation. Moreover, deep learning models can be trained on large-scale datasets, enabling them to generalize well and handle diverse maize disease classes. In this paper, we propose a deep learning-based framework for maize disease detection, leveraging the power of CNNs to automatically learn and extract meaningful representations from maize images, thereby providing an efficient and accurate solution for disease diagnosis and management in maize crops.

\section{Literature Survey}
Plant classification and disease detection have been active research areas in computer vision and machine learning. The classification of plants and detection of plant diseases have significant implications for agriculture, food security, and environmental sustainability. In recent years, deep learning has emerged as a powerful tool for plant classification and disease detection, with numerous studies demonstrating its effectiveness.

Plant disease detection has also been a popular research area in recent years. Deep learning-based methods have been shown to be effective in identifying plant diseases, even in complex scenarios. For example, Fuentes et al. (2017) \cite{fuentes2017} proposed a deep CNN model for the detection of citrus diseases. Their model achieved a classification accuracy of 97.33\%, outperforming traditional machine learning methods. In a similar study, Sladojevic et al. (2016) \cite{sladojevic2016} proposed a deep CNN model for the detection of grapevine diseases. Their model achieved an accuracy of 97.9\%, demonstrating the effectiveness of deep learning for plant disease detection.

There have also been several studies that have combined plant classification and disease detection. For example, Noh et al. (2019) \cite{noh2019} proposed a deep CNN model for the classification of rice diseases and pests. Their model achieved an accuracy of 94.58\% and was able to classify both the plant species and the disease or pest affecting the plant. Similarly, Singh et al. (2020) \cite{singh2020} proposed a deep CNN model for the identification of maize diseases. Their model achieved an accuracy of 99.12\% and was able to accurately detect multiple diseases affecting maize plants.
In conclusion, deep learning-based methods, particularly CNN architectures, have shown great promise in plant classification and disease detection tasks. The use of deep learning algorithms can significantly improve accuracy and outperform traditional machine learning methods. This research paper compares the efficiency of four CNN architectures, AlexNet\cite{krizhevsky2012imagenet}, GoogLeNet \cite{szegedy2015going}, EfficientNet  \cite{tan2019efficientnet}and Resnet \cite{he2016deep} and vision transformer  \cite{dosovitskiy2020image} to determine the best methodology for disease detection. The findings of this study can contribute to the development of effective techniques for disease detection in maize crops and aid in reducing the reliance on expensive and harmful chemical methods.
\subsection{Contributions}
In this paper, our contributions are the following:
\begin{itemize}
  \item We introduce a novel end to end pipeline for disease detection for maize crops with minimal human intervention.
  \item We collect and inference on a custom hand collected dataset that captures different phases of infected and non infected maize crops and leverage proper combination of data augmentation techniques to preserve features.
  \item We provide comparative study of different feature extractor architectures and conclude the GoogleNet as the most appropriate network for our task.
  \item Our pipeline achieves state of the art results on both PlantVillage Dataset and manually collected datasets with minimum human intervention.
\end{itemize}

\section{Methodology Used}

A convolutional neural network (CNN) is commonly used for image and video processing. The key feature of a CNN is the convolution operation, which enables the network to learn spatial features from the input image. In convolutional neural networks (CNNs), the input image is passed through a series of convolutional layers, where each layer consists of filters that progressively learn more intricate features. Subsequently, the output is passed through pooling layers that decrease the spatial dimensionality of the output. This makes them excellent for extraction of features from the data.

The convolution operation involves sliding a small window called a filter or kernel over the input image, which performs a dot product between the filter and the input pixels contained within the window, resulting in the generation of a single output value. The filter is then shifted to the next position in the input image, and the process is repeated, generating a 2D output map. The training process involves the collection of filter parameters through backpropagation, which serves to optimize the filter values in order to minimize the discrepancy between the predicted output and the ground truth labels.

The activation function plays an essential role in introducing nonlinearity to the network, allowing it to learn complex patterns in the input data. CNN's convolutional layers extract features from the input data by performing a linear transformation, where each neuron in the output is a weighted sum of the input pixels. Without nonlinearity, the network could only learn linear transformations of the input data, which is frequently insufficient for image and video processing tasks. The activation function is applied element-wise to the output of each neuron in the convolutional layer. Rectified Linear Unit (ReLU) is the most frequently used activation function in CNNs as it is found to be quicker to train. The three architectures used in this work are as follows:

\subsection{AlexNet}

In 2012, Alex Krizhevsky, Ilya Sutskever, and Geoffrey Hinton proposed the architecture of AlexNet, which is a convolutional neural network (CNN). The model emerged as the champion of the ImageNet Large Scale Visual Recognition Challenge (ILSVRC), surpassing the performance of the previous state-of-the-art models by a significant margin. AlexNet architecture (Fig. 1) consists of 5 convolutional layers, 3 max-pooling layers, 2 normalization layers, 2 fully connected layers, and 1 softmax layer. Each convolutional layer consists of convolutional filters and the nonlinear activation function ReLU.

\begin{figure}[htbp]
  \centering
  \includegraphics[width=0.4\textwidth]{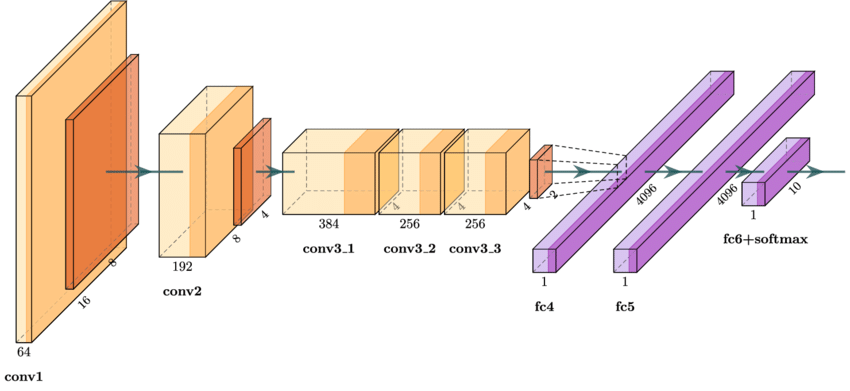}
  \caption{AlexNet Architecture}
\end{figure}

The input to the network is a 227x227 RGB image, which is first preprocessed by subtracting the mean RGB value of the training set, and the final output is a probability distribution over the 1000 different image categories. In addition, the architecture employs several regularization techniques, including dropout and data augmentation, to prevent overfitting and enhance generalization performance.

\subsection{GoogleNet}

GoogleNet, also known as Inception v1, is a deep convolutional neural network architecture developed in 2014 by Google researchers for large-scale image recognition tasks. The GoogleNet architecture (Fig. 2) consists of 22 layers with a unique design that incorporates multiple branches of convolutional layers of varying sizes, enabling the network to capture features at multiple scales. The most innovative aspect of the GoogleNet architecture is the inception module, 9 of which are used, which reduces the number of network parameters while increasing its depth. The inception module combines 1x1, 3x3, and 5x5 filter sizes in parallel, allowing the network to capture both local and global features. Each filter's output is concatenated along the depth dimension, resulting in an output with a greater depth. An important aspect of this architecture is the use of a global average pooling layer rather than entirely connected layers at the network's endpoints. This significantly reduces the number of network parameters and increases its robustness to input size variations.

\begin{figure}[htbp]
  \centering
  \includegraphics[width=0.4\textwidth]{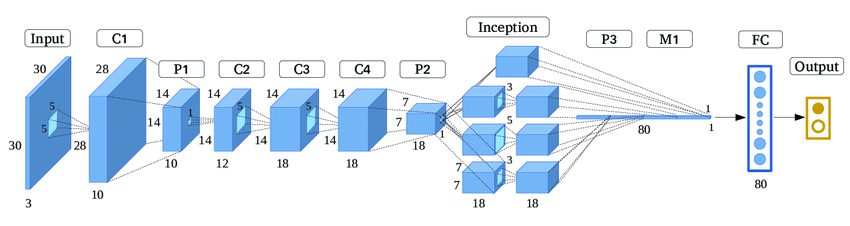}
  \caption{GoogleNet Architecture}
\end{figure}

\subsection{EfficientNet}

EfficientNet is an architecture and scaling method for convolutional neural networks that scales all dimensions of height, width, and depth uniformly using a compound coefficient. Instead of scaling width, depth, or resolution arbitrarily, compound scaling scales all three using a fixed set of scaling coefficients, as shown in Fig. 4. This distinguishes it from the other models examined. EfficientNet is based on the baseline network created through the neural architecture search utilizing the AutoML MNAS framework. The network is optimized for maximum precision but is penalized if it is computationally intensive. It is also penalized for slow inference time when the network takes a lot of time to make predictions.

\begin{figure}[htbp]
  \centering
  \includegraphics[width=0.4\textwidth]{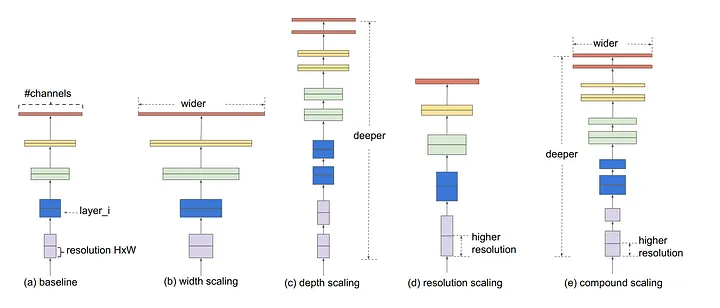}
  \caption{EfficientNet Architecture}
\end{figure}

The total number of layers in EfficientNet-B0 is 237       which are divided into 5 modules. These modules further combine to form sub-blocks, which then combine to form blocks. 
\begin{figure}[htbp]
  \centering
  \includegraphics[width=0.4\textwidth]{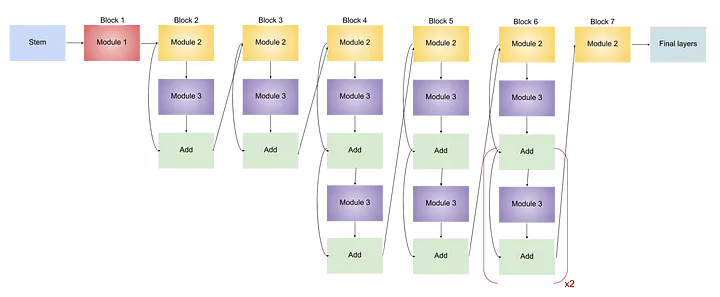}
  \caption{EfficientNet Architecture}
\end{figure}

\subsection{ResNet}

ResNet is a convolutional neural network (CNN) that uses "skip connections" to solve the vanishing gradient problem, enabling the creation of much larger neural network models. The process of backpropagation is used to train neural networks. This method employs gradient descent, shifting the loss function downward and determining its minimum weights. If multiple layers are employed, repeated multiplications reduce the gradient until it "disappears," and performance reaches a plateau or degrades with each additional layer. This is the problem of vanishing gradients. ResNet stacks multiple identity mappings (convolutional layers that initially do nothing), bypasses these layers, and reuses the activations from the previous layer. Skipping accelerates initial training by reducing the number of network layers. Then, when the network is retrained, all layers are expanded, and the residual parts of the network are permitted to explore a larger portion of the input image's feature space. Most ResNet models skip two or three layers at a time, with batch normalization and nonlinearity in between.
 \begin{figure}[htbp]
  \centering
  \includegraphics[width=0.4\textwidth]{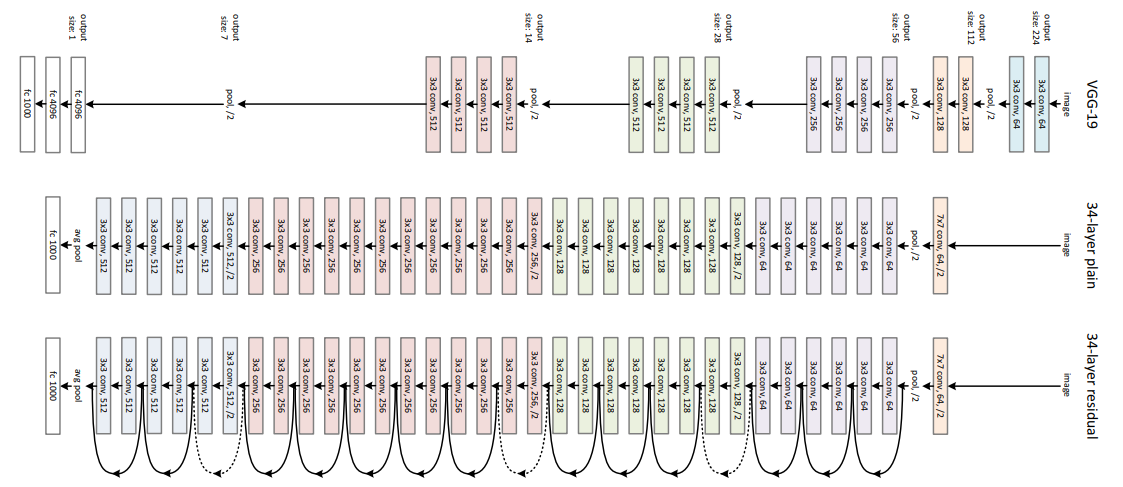}
  \caption{EfficientNet Architecture}
\end{figure}
\subsection{Vision Transformer}

Vision Transformer is a deep learning architecture for image classification introduced by Dosovitskiy et al. in a 2020 paper. It is based on the Transformer architecture, which was originally created for tasks involving natural language processing. Each transformer block in the Vision Transformer architecture includes a multi-head self-attention mechanism and a position-wise completely connected feedforward network. Patches extracted from the input image are flattened and projected into a lower-dimensional embedding space using a linear projection layer as input to the Vision Transformer. The self-attention mechanism of the Transformer block enables the network to model global dependencies between different regions, enabling it to capture spatial relationships between various image components. The position-wise feedforward network is used to introduce nonlinearity into the network and transform the features between layers. An important innovation of the Vision Transformer architecture is the inclusion of a "class token" to the input embedding to represent the overall class of the input image. This permits the network to perform classification duties without the need for additional fully connected layers at the network's endpoint.

\begin{figure}[htbp]
  \centering
  \includegraphics[width=0.4\textwidth]{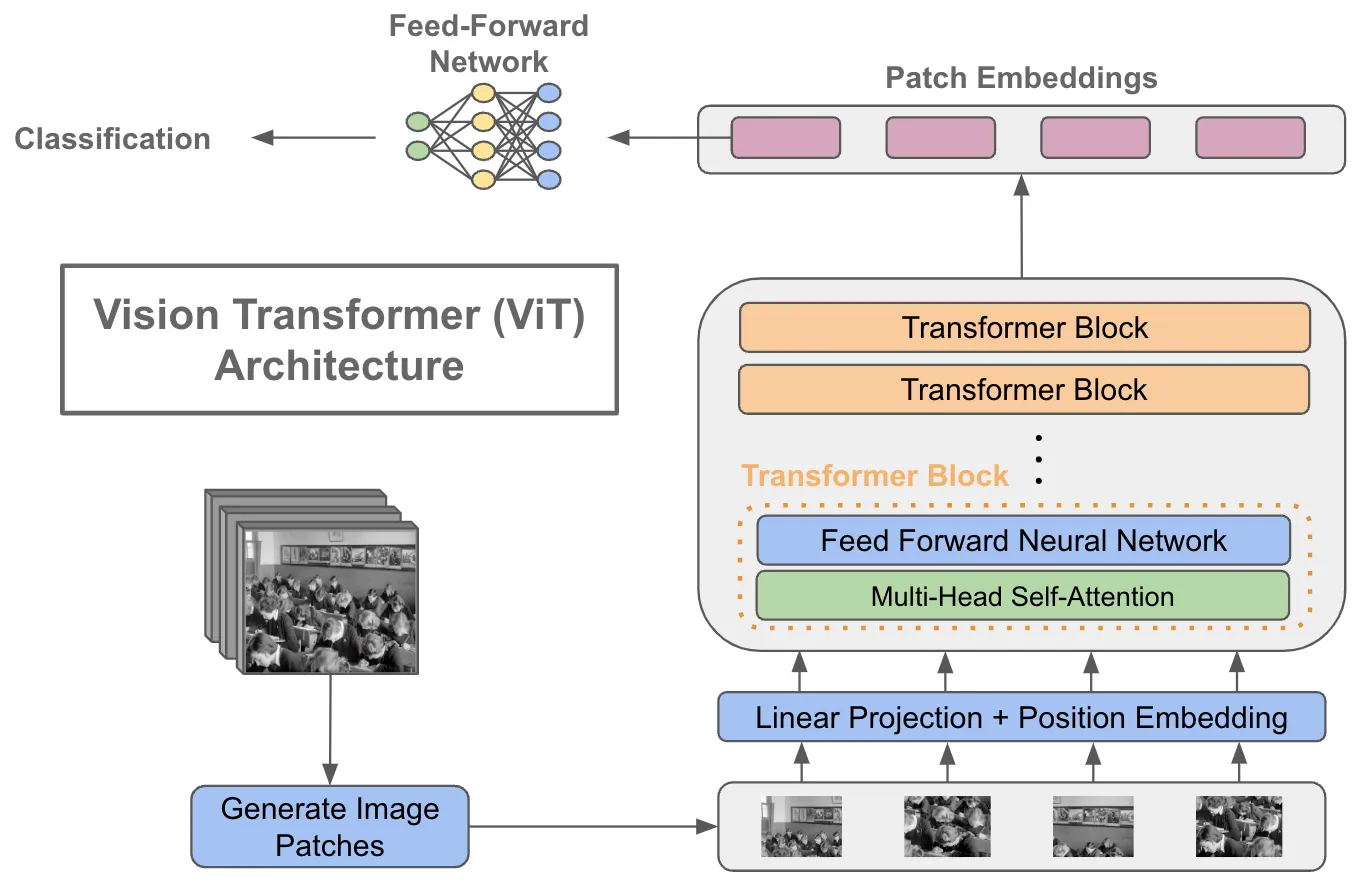}
  \caption{Transformer Architecture}
\end{figure}

We compare these networks for feature extraction and employ an ensemble \cite{dietterich2000ensemble} approach for the final output. Ensemble learning combines the predictions of multiple models to make a collective decision, often resulting in improved accuracy and robustness. We use Gradient Boosted Model called XGBoost \cite{chen2016xgboost} that iteratively trains weak models, typically decision trees, in a sequential manner, with each subsequent model attempting to correct the mistakes of the previous model. The final prediction is made by aggregating the predictions of all the weak models..
We further use triple loss \cite{schroff2015facenet} as a metric learning technique that aims to enhance the discriminative power of the learned feature representations. By incorporating triple loss into our ensemble, we enable the models to learn more effective feature embeddings that capture the underlying structure of the data and facilitate better discrimination between classes.

The triple loss is defined by the anchor, positive, and negative samples, with the objective of minimizing the distance between the anchor and positive samples, while maximizing the distance between the anchor and negative samples. This encourages the models to pull together samples from the same class and push apart samples from different classes, effectively enhancing inter-class separability. By utilizing triple loss in the ensemble learning process, we aim to enhance the ensemble's capability to distinguish between different classes and improve the overall classification accuracy.

\section{The Dataset}

The dataset (Fig. 7) used for this work is a custom maize dataset. It has 4,251 images divided into 4 categories, 3 of which are types of diseases (blight, common rust, and gray leaf spot), and one category containing images of healthy leaves. To create the custom collected maize dataset, high-resolution images of maize plants were acquired using a digital camera with a resolution of 12 megapixels. The imaging process was performed under controlled environmental conditions, with considerations made for lighting and weather conditions. Images of maize plants were captured at different growth stages, from seedlings to mature plants, and were collected from different locations in different regions. Additionally, the dataset was created to include multiple varieties of maize plants, each with varying levels of resistance to common diseases. We acquire these images through drones carrying multispectral sensors that with the use of vegetation indices detect distressed regions on the field, cutting down the manual effort from sampling from every tree to a minimum percentage of them. We then use instance segmentation techniques to identify what parts of the field we wish to collect leaf samples from and acquire images from those parts of field.

The acquired images were labeled manually by domain experts using a standard labeling protocol of identifying and marking images of maize plants affected by various diseases, including leaf blight, gray leaf spot, and common rust. Each image was labeled with the corresponding disease, and the labeling process was repeated multiple times by different experts to ensure the accuracy of the labels. We finally acquire 1276 images of commun rust, 634 images of Gray Leaf Spot, 1126 images of Blight and 1215 Healthy images.

Image classification models, particularly in the domain of disease detection in plants, greatly benefit from the application of data augmentation \cite{ghorbani2020plant} techniques. These techniques play a vital role in enhancing the performance and robustness of the models by introducing variations and expanding the diversity of the training dataset. Previous studies in this field often overlooked the incorporation of data augmentation, limiting the models' ability to generalize to real-world scenarios.

Traditional approaches to disease detection in plants relied on small and homogeneous datasets, resulting in models that struggled to capture the full complexity and variations present in plant diseases. However, in our study, we recognize the importance of data augmentation techniques as a means to address this limitation. By employing a comprehensive set of augmentation strategies, including rotation, flipping, zooming, scaling, cropping, translation, noise injection, color jittering, elastic transformation, occlusion, and channel shifting, we ensure that our model learns from a diverse range of images, capturing the inherent variability of plant diseases.

These data augmentation techniques serve to introduce variations in the dataset, simulating real-world scenarios such as changes in scale, orientation, viewpoint, lighting conditions, and occlusions. Rotation and flipping enable the model to learn from images with different orientations, while zooming and scaling account for the varied sizes at which plant diseases may occur. Translation introduces positional shifts, mimicking different positions within the frame. Noise injection and color jittering simulate real-world variations in image attributes. Elastic transformation models small distortions that can occur due to various factors, while occlusion provides the model with exposure to partially occluded instances. Channel shifting alters the color distribution, making the model more robust to variations in color.

By augmenting the dataset using these techniques, our study ensures that our model is exposed to a more diverse and representative set of training examples. This diverse training set allows the model to learn a more comprehensive set of features and patterns, enabling it to better generalize and accurately classify plant diseases. As a result, our study demonstrates improved performance compared to earlier approaches, achieving higher accuracy and robustness in disease detection tasks.

 \begin{figure}[h]
  \centering
  \includegraphics[width=0.4\textwidth]{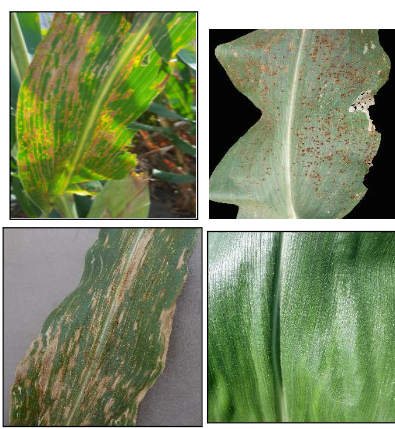}
  \caption{Dataset Snapshot }
\end{figure}

\subsection{Results}

\textbf{Alexnet}

The model was trained for 30 epochs, saved, and then run again for 15 epochs. The batch size used was 4, and the learning rate was 1e-5. The final results obtained are as follows:
\begin{figure}[htbp]
  \centering
  \includegraphics[width=0.4\textwidth]{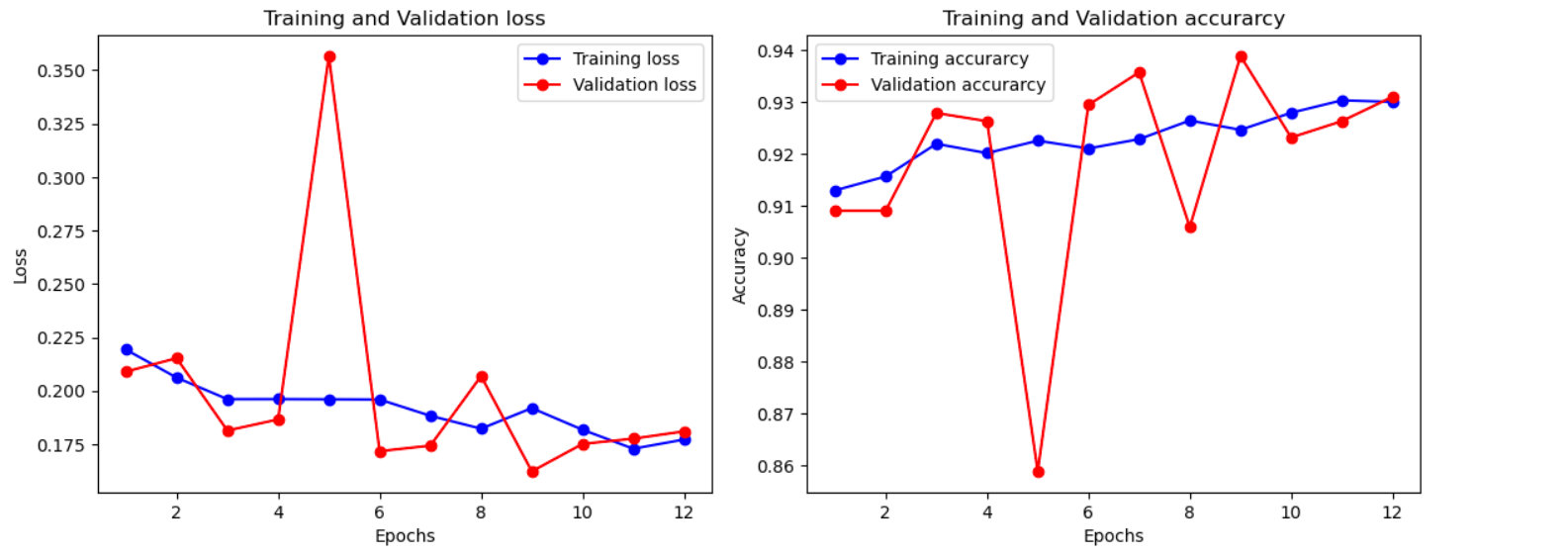}
  \caption{AlexNet Results }
\end{figure}

\bigskip

\textbf{GoogleNet}

The model was trained and tested by running it with 50 epochs on a batch size of 16. The learning rate was set at 0.001, and momentum was set at 0.9. To prevent overfitting, the images were randomly rotated horizontally while loading into the train loader. With each epoch, a train cycle and a test cycle were run, resulting in the loss function and accuracy of the model. The model was trained for 50 epochs. The final accuracy of the testing set received was 99.90\%, and the best accuracy achieved was 100\%, making it the most ideal architecture in terms of accuracy.
\begin{figure}[htbp]
  \centering
  \includegraphics[width=0.4\textwidth]{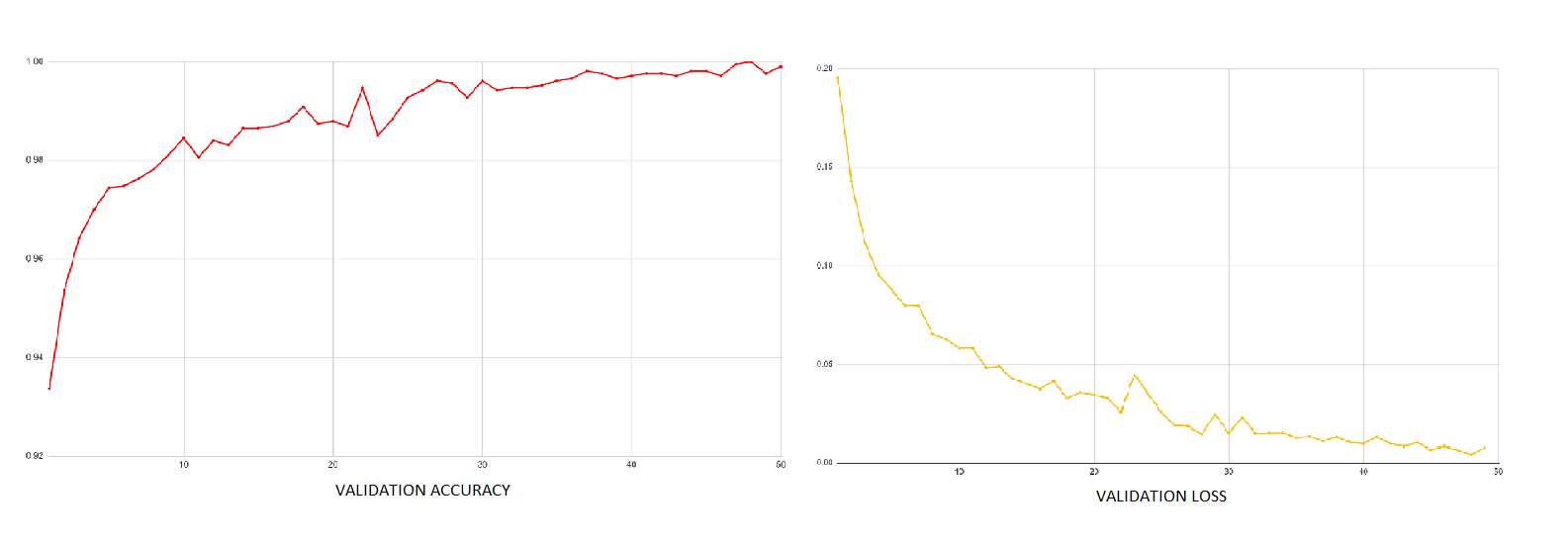}
  \caption{GoogleNet Results }
\end{figure}

\bigskip

\textbf{Vision Transformer}

The transformer was run for 5 epochs, which took 15 hours. The batch size used was 32, and the learning rate was 2e-5. The final accuracy on the testing set received was 99.23
\begin{figure}[h]
  \centering
  \includegraphics[width=0.4\textwidth]{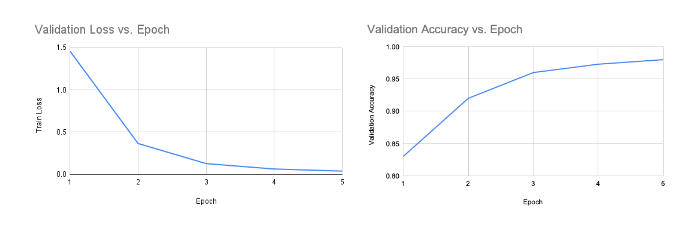}
  \caption{Vision Transformer Results }
\end{figure}

\bigskip

\textbf{EfficientNet}

The model was trained for 25 epochs. The final and best accuracy on the testing set received was 99.95\%. This value stagnated for 5 epochs. The batch size was set at 32, the learning rate at 3e-3, and the momentum at 0.9. The number of classes was 4. The final and best accuracy on the testing set received was 97.5\%. This value stagnated for 3 epochs.

\begin{figure}[htbp]
  \centering
  \includegraphics[width=0.4\textwidth]{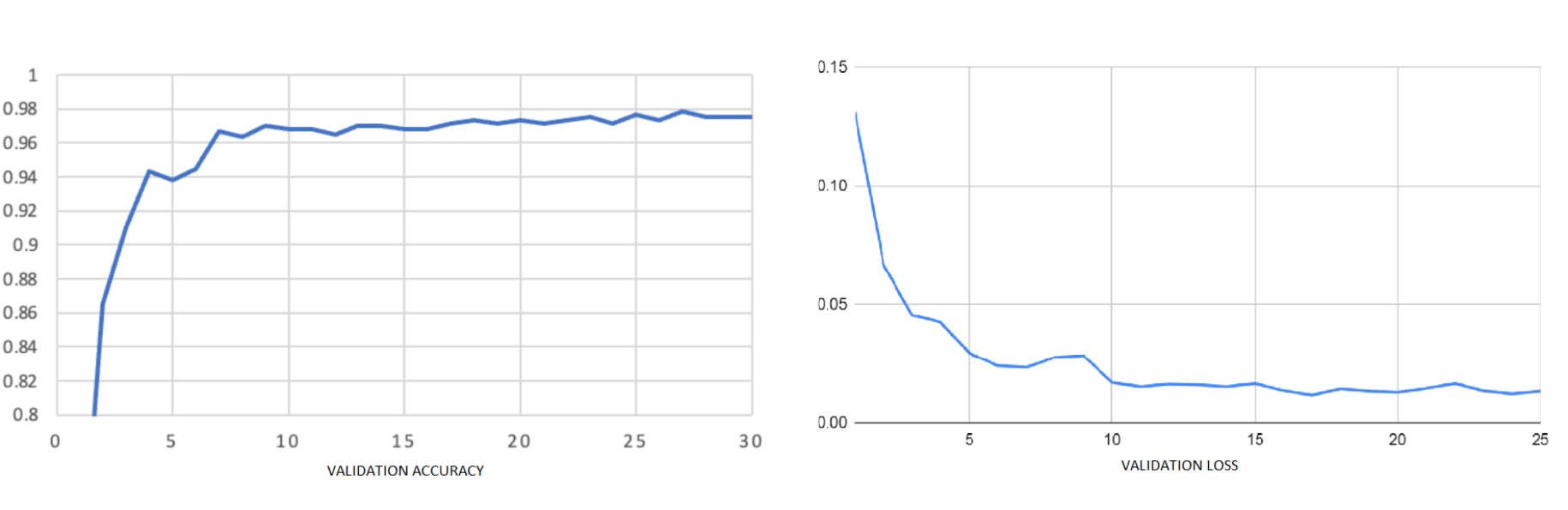}
  \caption{EfficientNet Results }
\end{figure}

\bigskip

\textbf{ResNet-50}

The model was trained for 35 epochs. The batch size used was 16, and the learning rate was 1e-3. The final validation accuracy obtained was 97.66\%.
\begin{figure}[htbp]
  \centering
  \includegraphics[width=0.4\textwidth]{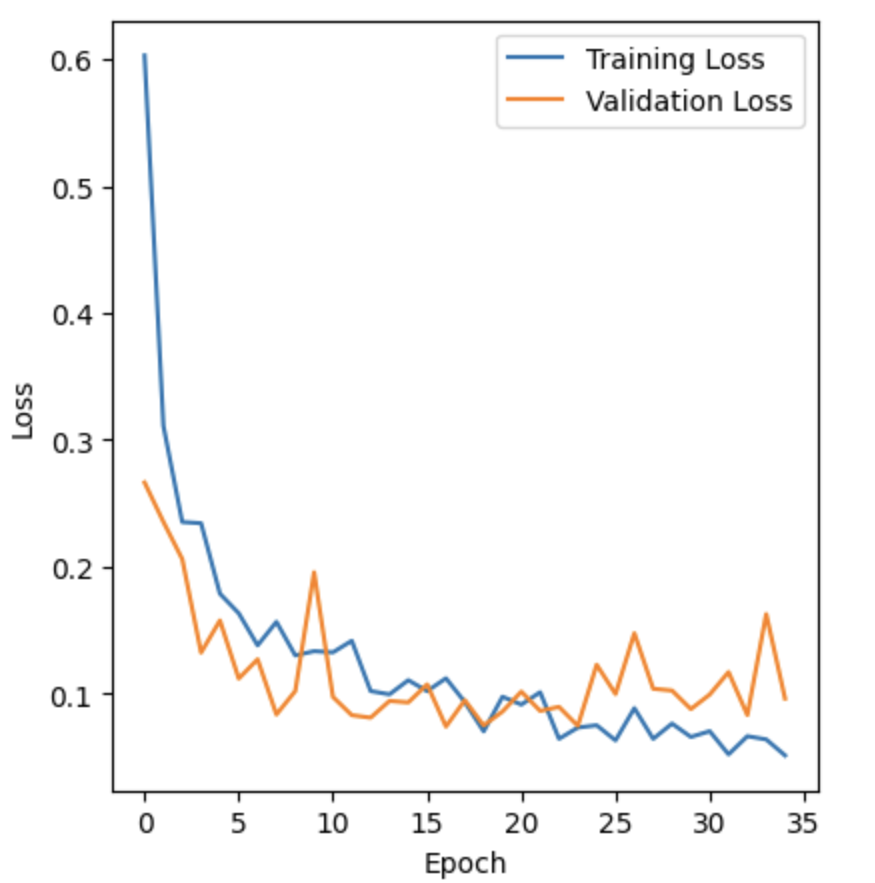}
  \caption{Resnet Results }
\end{figure}

\begin{table}[htbp]
\centering
\caption{Overall Comparison of Model Performances}
\begin{tabular}{|l|c|c|}
\hline
\textbf{Model Name} & \textbf{Test Accuracy (\%)} & \textbf{Model Loss} \\ \hline
AlexNet & 93.5 & 0.175 \\ \hline
\textbf{GoogleNet} & \textbf{99.8} & 0.0637 \\ \hline
EfficientNet & 97.50 & 0.0138 \\ \hline
ResNet & 97.66 & 0.0961 \\ \hline
Vision Transformer & 99.23 & 0.023 \\ \hline
\end{tabular}
\end{table}

\section{Conclusions}

AMaizeD: An End to End Pipeline for Automatic Maize Disease Detection addresses the challenges faced by deep learning models in automatic disease detection, particularly when applied to real-world images. We have demonstrated that the GoogleNet architecture serves as a powerful feature extractor, exhibiting superior performance in classifying diseases in agricultural crops. Moreover, to further enhance the accuracy and robustness of our disease detection system, we employed ensemble learning with XGBoost. This combination not only achieved state-of-the-art results but also showcased the potential for obtaining exceptional performance with minimal human intervention. The utilization of ensemble learning techniques allowed us to leverage the collective decisions of multiple models, providing a more comprehensive and accurate classification outcome. Notably, while newer architectures such as the Vision Transformer show promise in matching the accuracies of conventional CNN models, their computationally intensive nature and data requirements make them less practical for real-world applications. Overall, our study highlights the efficacy of combining the GoogleNet feature extractor with ensemble learning using XGBoost, yielding impressive results and reducing the need for extensive human intervention in the disease detection of maize crops process. 
\section{Acknowledgements}
This work was supported by the Short Term Project Scheme of Technology Development Programme (DRISHTI CPS/TDP-ST/SL/2022/003) from IITI DRISHTI CPS Foundation.

\bibliography{egbib}

\end{document}